\pdfoutput=1
\documentclass[sigconf,nonacm]{acmart}
\usepackage{amsmath}
\usepackage{array}
\usepackage{booktabs}
\usepackage{enumitem}
\usepackage{tabularx}
\usepackage{makecell}
\usepackage{algorithm}
\usepackage{algorithmic}
\usepackage{colortbl}
\usepackage{siunitx}
\definecolor{linkplace}{HTML}{159A8C}  %
\definecolor{linkinfo}{HTML}{2F6DB5}   %
\definecolor{linkimpl}{HTML}{D07A1F}   %
\newcommand{\linkchip}[1]{\textcolor{#1}{\rule[0.05ex]{0.6em}{0.6em}}\,}
\definecolor{tabband}{HTML}{EBEFF4}
\definecolor{tabgroup}{HTML}{F3F5F8}
\definecolor{tabours}{HTML}{E1F1EE}
\definecolor{tabsub}{HTML}{5B6773}
\definecolor{tabhair}{HTML}{CBD2D9}
\newcommand{\tablestyle}{\setlength{\aboverulesep}{0pt}\setlength{\belowrulesep}{0pt}\renewcommand{\arraystretch}{1.18}}
\newcommand{\tabnote}[1]{\par\vspace{3pt}{\footnotesize\color{tabsub}#1\par}}

\title[Order-Flow Interventions After Dealer Adaptation]{When and How Order-Flow Interventions Work \texorpdfstring{\\}{}After Dealers Adapt}
\titlenote{Accepted at the 7th ACM International Conference on AI in Finance (ICAIF 2026).}

\author{Guanghao Wu}
\authornote{Both authors contributed equally to this research.}
\orcid{0009-0009-8882-6512}
\affiliation{%
  \institution{University of Technology Sydney}
  \city{Sydney}
  \state{NSW}
  \country{Australia}}
\email{guanghao.wu@student.uts.edu.au}

\author{Zhuo Cai}
\authornotemark[2]
\orcid{0009-0006-1649-4094}
\affiliation{%
  \institution{University of Technology Sydney}
  \city{Sydney}
  \state{NSW}
  \country{Australia}}
\email{zhuo.cai@student.uts.edu.au}

\author{Zhidong Li}
\orcid{0000-0002-0784-157X}
\affiliation{%
  \institution{University of Technology Sydney}
  \city{Sydney}
  \state{NSW}
  \country{Australia}}
\email{zhidong.li@uts.edu.au}

\author{Shoujin Wang}
\authornote{Corresponding author.}
\orcid{0000-0003-1133-9379}
\affiliation{%
  \institution{University of Technology Sydney}
  \city{Sydney}
  \state{NSW}
  \country{Australia}}
\email{shoujin.wang@uts.edu.au}

\begin{document}

\begin{abstract}
Learning dealer agents can settle on persistently wide quotes, raising what customers pay to trade. A venue can route more orders to tighter quotes, but a wide quote may reflect real risk, and once dealers adapt, an end-to-end comparison cannot show why a rule works: because it intervenes more, picks better moments, is announced in advance, or rewards the right quote. We propose \textbf{SOFA} (\emph{State-aligned Order-Flow Allocation}), a public order-flow contract with three parts: a \emph{State Scorer} ranks market states from lagged public data, a \emph{Commitment Scheduler} announces a budgeted intervention strength before dealers quote, and a \emph{Quote-to-Flow Allocator} shifts a bounded share of fixed customer demand toward tighter quotes. Our paired re-adaptation protocol gives competing rules the same intervention strengths, only re-timed, lets dealers re-adapt to each rule from a shared checkpoint, and breaks one link at a time. Across 20 independently trained dealer populations in an ABIDES market, concentrating the intervention in states that the score ranks as low-risk lowers low-risk execution cost relative to random timing in every population, by 0.040 tick on average, while a stale announcement or a broken quote-to-flow link raises this cost. In the primary setting the gain comes at the expense of high-risk trades, whose costs rise, so state alignment pays off only for a venue that weighs low-risk costs more than 1.38 times as heavily as high-risk costs. Our code is available at \href{https://github.com/moderncavemann/state-aligned-order-flow-contracts}{\mbox{\nolinkurl{https://github.com/}}\allowbreak\mbox{\nolinkurl{moderncavemann/state-aligned-order-flow-contracts}}}.
\end{abstract}



\maketitle

\section{Introduction}
\label{sec:introduction}

In dealer markets, market makers (dealers) quote the prices at which they stand ready to buy and sell, and the gap between the two---the bid--ask spread---is what customers pay to trade. Independent reinforcement-learning agents can sustain high prices in repeated-pricing settings \cite{calvano2020artificial,klein2021autonomous}, and learning-based dealer models can likewise settle into persistently wide quotes \cite{cont2024dynamics,colliard2026algorithmic,wang2025multi}. Yet a wide spread is not, by itself, evidence of strategic behavior: it can also reflect volatility, adverse selection (trading against better-informed counterparties), or inventory risk (holding risky positions) \cite{glosten1985bid,avellaneda2008high}. A venue that intervenes whenever spreads widen can suppress legitimate risk compensation as readily as strategic margins.

\begin{figure}[h]
\centering
\includegraphics{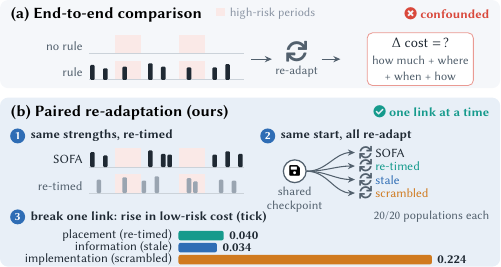}
\caption{Judging a rule after dealers adapt. (a) End-to-end comparisons change how much, where, when, and how the venue intervenes at once; (b) paired re-adaptation fixes the strengths, re-adapts every arm from one checkpoint, and breaks one link per control (sizes also reflect how disruptive each control is).}
\label{fig:teaser}
\Description{Two stacked panels. Panel a shows a strip of intervention strengths over time for a rule and an empty strip for no rule; after dealers re-adapt, the cost difference mixes how much, where, when and how. Panel b shows two strips with the same strength bars in different time slots, the SOFA strip avoiding shaded high-risk periods; a shared checkpoint fans out to SOFA, re-timed, stale and scrambled arms that each re-adapt; and three bars of the rise in low-risk cost when one link is broken: placement 0.040, information 0.034, implementation 0.224 tick, each in 20 of 20 populations.}
\end{figure}

We ask what such a venue can actually do. It cannot inspect dealer objectives, parameters, or learning updates: to the venue, dealers are black boxes that keep learning. What the venue does control is how customer orders reach dealers: it observes lagged public market data, announces allocation terms before dealers quote, and executes those terms after quotes arrive while keeping total customer demand fixed. This is a black-box multi-agent reinforcement learning (MARL) governance problem: the venue can change how learning agents earn customer flow, but not how they learn. We call such a rule a \emph{contract} in this implementation sense---a public commitment made before dealers quote and enforced mechanically---rather than a bilateral legal contract or an equilibrium guarantee.

Designing such a contract raises three challenges. \textbf{CH1 (where): risk-confounded opportunity.} From lagged public data alone, the venue must find the states in which wide spreads are least explained by risk; a rule that simply reacts to wide spreads is confounded, because spreads also widen when risk rises. \textbf{CH2 (when and how much): budgeted commitment before quoting.} Opportunities are scarce, continuous intervention inflates dealers' inventory penalties (Section~\ref{sec:sensitivity}), and terms announced after quoting cannot shape the quote. \textbf{CH3 (how): demand-conserving implementation.} Unlike an incentive designer, the venue cannot transfer rewards; it must keep total demand fixed while tying the promised flow advantage to the quote that earned it. The challenges are linked: well-placed intervention is wasted if dealers cannot see the current commitment, and an announcement is wasted if the allocation ignores each dealer's own quote.

Evaluating such a contract raises a fourth challenge. \textbf{CH4 (how to judge): identification after adaptation.} Once dealers adapt to a new rule, which part of the rule caused the change? An end-to-end comparison changes at once how much and where the venue intervenes, what dealers know before quoting, and how quotes map to flow (Figure~\ref{fig:teaser}a). A standard ablation that removes a component also changes how much the venue intervenes, mixing the amount of intervention with the mechanism, while a test with frozen dealer policies captures only the mechanical effect before dealers adapt. The question is therefore not whether a favorable final outcome occurs, but which links still matter after dealers have adapted. Answering it requires a design that fixes the amount of intervention, lets every compared rule re-adapt, and changes one link at a time.

We address CH1--CH3 with \textbf{SOFA} (\emph{State-aligned Order-Flow Allocation}), a public order-flow contract whose three components map one-to-one onto the design challenges. The \emph{State Scorer} turns lagged public volatility and absolute returns into a frozen score that ranks states by how little risk would explain a wide spread (CH1). The \emph{Commitment Scheduler} converts the score into a paced intervention strength and announces it before dealers quote (CH2). The \emph{Quote-to-Flow Allocator} then shifts a bounded share of the fixed customer demand toward tighter quotes, so the promised advantage goes to the quote that earns it (CH3).

We address CH4 with a \emph{paired re-adaptation protocol} (Figure~\ref{fig:teaser}b). On common market paths, SOFA and its placement controls (each complete rule under comparison is an \emph{arm}) receive exactly the same set of intervention strengths, placed at different times, and every arm re-adapts from the same trained checkpoint with the same training budget. Targeted controls then break one link at a time: a stale announcement changes what dealers see before quoting, and identity scrambling gives each dealer the flow earned by another dealer's quote. We repeat the protocol on 20 independently trained dealer populations, so each conclusion rests on population-level evidence rather than on a single training run.

The paper makes three contributions:
\begin{enumerate}[leftmargin=*,label=(\arabic*),itemsep=0pt,topsep=2pt]
\item \textbf{A paired re-adaptation protocol} that holds the set of announced strengths fixed on every path and re-adapts every arm from a shared checkpoint, so outcome differences can be attributed to placement, information, or implementation rather than to intervening more or to dealers that have not yet adapted.
\item \textbf{SOFA, a traceable order-flow contract} that needs no access to dealer rewards or gradients and logs a replayable chain from state to score, announcement, quote, allocation, and outcome.
\item \textbf{Evidence on what works after adaptation, and at whose cost:} a consistent gain from state-aligned placement, an advantage of current over stale announcements, the largest tested loss when the quote-to-flow link is broken, the limits of transfer across learners and market settings, and an account of who pays for the targeted gain and when the trade is worth making.
\end{enumerate}

\section{Related Work}
\label{sec:related-work}

\textbf{Algorithmic coordination and dealer risk.} Algorithmic-pricing research asks whether learning agents sustain supracompetitive prices, through which punishment-style strategies, and under which learning dynamics \cite{calvano2020artificial,klein2021autonomous,hartline2025regulation,bichler2026convergence}. Dealer microstructure instead explains wide spreads as compensation for inventory risk and adverse selection \cite{glosten1985bid,avellaneda2008high}, and recent models join the two views \cite{cont2024study,colliard2026algorithmic,wang2025multi}. Because a wide spread may reflect strategy or risk, we target a conditional intervention rather than a collusion label.

\textbf{Market making and order allocation.} Market-making research derives dealer quotes from inventory and transaction risk and optimizes them with deep or adversarial reinforcement learning \cite{ho1981optimal,gueant2013dealing,gueant2019deep,spooner2020robust}, while multi-agent dealer models let competition among dealers emerge from learning itself \cite{ganesh2019reinforcement,cont2024dynamics,wang2025multi}. Optimal-order-placement work asks how a trader should route orders against a given market \cite{cont2017optimal}. We take the opposite, venue-side view: the venue commits how a fixed customer flow responds to public state and submitted quotes, and dealers adapt to that rule; our contribution is an evaluation design for such rules, not a new dealer objective or quoting algorithm.

\textbf{Steering adaptive agents.} Platform rules and learned buyer commitments can reshape seller responses \cite{johnson2023platform,brero2022learning,banerjee2026interpretingcounteringcollusiondeeplearning}, and MARL incentive designers can transfer rewards directly \cite{yang2020learning}. Our venue cannot transfer rewards; it can only commit an allocation rule before dealers quote, which we freeze and test link by link after every arm re-adapts.

\textbf{Financial MARL evaluation.} ABIDES supports message-level counterfactuals \cite{byrd2020abides}, ABIDES-MARL studies strategic learning \cite{cheridito2025abides}, and JaxMARL-HFT emphasizes GPU-scale historical-data learning \cite{mohl2025jaxmarl}. These frameworks make large-scale training feasible but do not attribute an outcome change to one link of a fixed intervention after agents adapt, which we do on common ABIDES event paths. MAPPO and IPPO provide distinct learners \cite{yu2022surprising,de2020independent}, and MARL evaluation guidance \cite{agarwal2021deep,gorsane2022towards} motivates our inference across independently trained populations.

\section{Problem Formulation}
\label{sec:problem-formulation}

\subsection{Market Model}

Let $\mathcal N=\{1,\ldots,N\}$ denote the dealer set and $\mathcal T=\{1,\ldots,T\}$ the sequence of market decisions. At decision $t\in\mathcal T$, $m_t$ is the public pre-shock reference price. Dealer $i\in\mathcal N$ submits a symmetric half-spread $s_{i,t}\in\mathcal S$---the distance of its quotes from the reference price, so a larger half-spread means customers trade at worse prices---and $\mathbf s_t=(s_{1,t},\ldots,s_{N,t})$ denotes the submitted quote profile. The venue routes a fixed customer quantity $Q$ using allocation weights $\mathbf w_t=(w_{1,t},\ldots,w_{N,t})$ satisfying $w_{i,t}\geq0$ and $\sum_i w_{i,t}=1$.

The public reference evolves as $m_{t+1}=m_t+\epsilon_t$. A simulator-internal regime $z_t\in\{\mathrm L,\mathrm H\}$ changes the conditional scale of $\epsilon_t$, while the informed-arrival probability remains fixed. The regime therefore changes the magnitude of adverse selection, not the frequency of informed flow. It drives the simulator, labels the offline calibration data, and defines the post-hoc evaluation subgroups, but is unavailable to the venue at runtime.

For dealer $i$, $C_{i,t}$ and $I_{i,t}$ denote cash and inventory, and $W_{i,t}=C_{i,t}+I_{i,t}m_t$ is wealth marked to the reference price. Each dealer's learning reward penalizes inventory:
\begin{equation}
r_{i,t}=W_{i,t+1}-W_{i,t}-\lambda_I I_{i,t+1}^{2}.
\label{eq:reward}
\end{equation}
The venue's runtime information is the lagged public history $\mathcal H^{\mathrm{pub}}_{t-1}$. It computes a public-state score $q_t$, converts it into an announced strength $u_t$ using scheduler state $B_{t-1}$, and allocates customer flow after observing the current quote profile:
\begin{equation}
q_t=M_1(\mathcal H^{\mathrm{pub}}_{t-1}),\qquad
u_t=M_2(q_t,B_{t-1}),\qquad
\mathbf w_t=M_3(u_t,\mathbf s_t).
\label{eq:information-order}
\end{equation}
Thus the State Scorer ($M_1$) has no latent-state or current-quote input, dealers observe $u_t$ before choosing $\mathbf s_t$, and the Quote-to-Flow Allocator ($M_3$) uses only $u_t$ and the submitted quotes.

\subsection{Venue Objective}

The venue targets customer execution in states where wide spreads are less justified by shock-magnitude risk. Let $\mathcal E_t$ be the set of customer executions at decision $t$; for execution $e\in\mathcal E_t$, let $q_e$, $p_e$, and $m_t$ denote filled quantity, execution price, and contemporaneous reference price. The prespecified primary endpoint, which we call the \emph{low-risk execution cost}, is the quantity-weighted customer execution half-spread in the low-risk regime:
\begin{equation}
Y_{\mathrm L}
=
\frac{
\sum_t\mathbf1[z_t=\mathrm L]
\sum_{e\in\mathcal E_t}q_e|p_e-m_t|
}{
\sum_t\mathbf1[z_t=\mathrm L]
\sum_{e\in\mathcal E_t}q_e
}.
\label{eq:primary-endpoint}
\end{equation}
Lower $Y_{\mathrm L}$ means cheaper customer execution; spreads, and all spread effects below, are measured in ticks, the minimum price increment. We also report where the cost falls and how the market responds---high-risk and pooled ($Y_{\mathrm{all}}$) execution cost, total customer cost in tick-units (the per-path sum of $q_e|p_e-m_t|$), dealer reward, inventory dispersion, routing intensity, and flow concentration---rather than collapsing them into a single welfare number; targeting $Y_{\mathrm L}$ is a choice of venue objective, not a welfare claim (Section~\ref{sec:worth}). The task is to test whether, under an identical set of strengths on each path, giving the stronger interventions to higher-scored states lowers $Y_{\mathrm L}$ after dealers adapt.

\subsection{Evaluation Targets}

Let $\mathbf u=(u_1,\ldots,u_T)$ denote the realized pathwise strength schedule. State-aligned, exact-random, circular-shift, and periodic schedules preserve the complete multiset of entries in $\mathbf u$. They consequently preserve the activation count and cumulative strength,
\begin{equation}
A(\mathbf u)=\sum_t\mathbf1[u_t>0],
\qquad
S(\mathbf u)=\sum_tu_t,
\label{eq:budget}
\end{equation}
as well as every individual nonzero strength.

For population $k$, each outcome is first averaged over its $R$ common evaluation paths. The three primary contrasts are
\begin{equation}
\begin{aligned}
\Delta_{\mathrm{place}}
&=
\mathbb E_k[
\bar Y^{\mathrm{random}}_{k,\mathrm L}
-
\bar Y^{\mathrm{aligned}}_{k,\mathrm L}
],\\
\Delta_{\mathrm{info}}
&=
\mathbb E_k[
\bar Y^{\mathrm{stale}}_{k,\mathrm L}
-
\bar Y^{\mathrm{current}}_{k,\mathrm L}
],\\
\Delta_{\mathrm{impl}}
&=
\mathbb E_k[
\bar Y^{\mathrm{scrambled}}_{k,\mathrm L}
-
\bar Y^{\mathrm{quote\text{-}aligned}}_{k,\mathrm L}
].
\end{aligned}
\label{eq:estimands}
\end{equation}
Positive values favor state alignment, the current announcement, and quote-aligned allocation, respectively; paired arms differ in exactly one link. Every arm re-adapts independently from the same population checkpoint, so these are total learned contrasts under a fixed simulator, learner, and training budget.

\section{Methodology}
\label{sec:methodology}

\begin{figure*}[t]
\centering
\includegraphics[width=\textwidth]{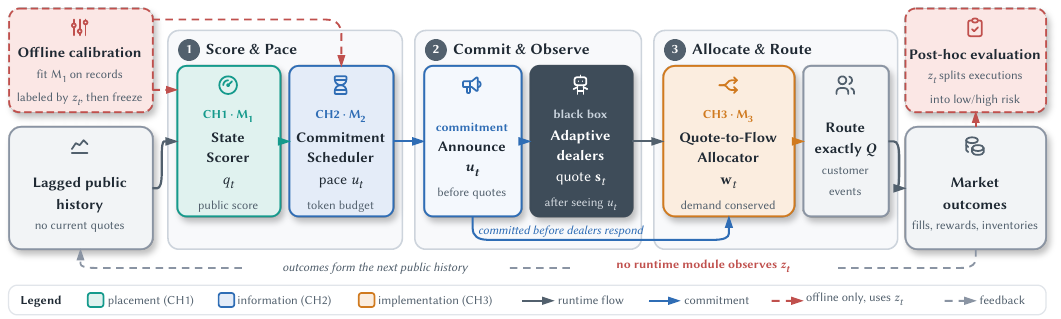}
\caption{\textbf{SOFA} at one market decision. (1) The State Scorer scores lagged public history and the Commitment Scheduler paces a token budget; (2) the strength $u_t$ is announced before dealers quote; (3) the Quote-to-Flow Allocator applies the committed $u_t$ to the submitted quotes. Colors mark the three links (CH1--CH3); the legend explains the line styles. The latent regime $z_t$ is used only offline.}
\label{fig:framework}
\Description{Three numbered stages between inputs on the left and outputs on the right. Stage 1, Score and Pace, holds the State Scorer and the Commitment Scheduler; stage 2, Commit and Observe, holds the announcement of the strength before quotes and the adaptive dealers, drawn as a black box that quotes after seeing it; stage 3, Allocate and Route, holds the Quote-to-Flow Allocator and the routing of exactly Q customer events. A blue line under stage 2 carries the committed strength to the allocator. Lagged public history feeds the scorer; market outcomes feed post-hoc evaluation and, along a dashed line at the bottom, the next public history. Offline calibration, drawn as a dashed box, feeds the scorer and the scheduler; the latent regime appears only in the two dashed offline boxes. Each box carries a small icon, and a legend along the bottom explains the link colors and line styles.}
\end{figure*}

\subsection{Overview}
Our method has two layers. The \emph{intervention layer} (Sections~\ref{sec:scorer}--\ref{sec:allocator}, Figure~\ref{fig:framework}, Algorithm~\ref{alg:runtime}) specifies what the venue does at each decision and answers CH1--CH3: the State Scorer decides where to intervene, the Commitment Scheduler decides when and how strongly, and the Quote-to-Flow Allocator decides how the promised advantage reaches quotes. The \emph{evaluation layer} (Section~\ref{sec:protocol}, Figure~\ref{fig:teaser}b) tests which of these parts still works after dealers adapt and answers CH4. All three runtime modules use only public or venue-side information, and every quantity they produce is logged.

\subsection{State Scorer: Where to Intervene}
\label{sec:scorer}

The State Scorer asks, from recent public data alone, whether this is a moment in which a wide spread would be hard to justify by risk. It maps lagged public history to a bounded score that ranks states before current quotes exist, using ten-step public volatility $v_t$ and the absolute lagged return $a_t$:
\begin{equation}
q_t=\sigma\!\left(b+\boldsymbol\beta^\top D^{-1}\left([v_t,a_t]^\top-\boldsymbol\mu\right)\right),\qquad q_t\in[0,1].
\label{eq:m1}
\end{equation}
The model is deliberately simple, which keeps the study focused on the intervention rather than on representation learning. It is fitted once on records without intervention, with the simulator's regime as the training label, and frozen before any evaluation; at runtime it reads neither the latent regime, current quotes, dealer identities, nor future outcomes. The score is therefore a public control index, not a live collusion probability. On held-out records it ranks the two regimes well (Table~\ref{tab:config}); it is not calibrated as a probability, which ranking does not require.

\subsection{Commitment Scheduler: When to Intervene}

The Commitment Scheduler turns the score into a commitment that is limited by a token budget and visible before dealers quote:
\begin{equation}
\begin{aligned}
\widetilde u_t&=q_t\mathbf1[q_t\geq\tau],&B_t^+&=\min\{B_{\max},B_{t-1}+\rho\},\\
u_t&=\widetilde u_t\mathbf1[B_t^+\geq1],&B_t&=B_t^+-\mathbf1[u_t>0].
\end{aligned}
\label{eq:m2}
\end{equation}
Pacing prevents continuous intervention, and announcing the strength before quoting lets dealers respond to the current commitment. In the primary setting the scheduler intervenes in about one decision in five. For a fixed set of strengths, giving the larger strengths to the states with the larger benefits is the static benchmark---the best arrangement if dealers did not react; the paired experiment tests whether the ordering still helps after dealers adapt to it.

\begin{algorithm}[t]
\caption{Runtime of the \textbf{SOFA} contract on one market path}
\label{alg:runtime}
\begin{algorithmic}[1]
\REQUIRE frozen $M_1$; scheduler parameters $\tau$, $\rho$, $B_{\max}$; allocator parameters $\kappa$, $\phi$; fixed customer demand $Q$
\STATE $B_0\gets0$
\FOR{$t=1,\ldots,T$}
\STATE Observe the lagged public history $\mathcal H^{\mathrm{pub}}_{t-1}$
\STATE Score the state: $q_t\gets M_1(\mathcal H^{\mathrm{pub}}_{t-1})$ \hfill Eq.~(\ref{eq:m1})
\STATE Pace the budget: $(u_t,B_t)\gets M_2(q_t,B_{t-1})$ \hfill Eq.~(\ref{eq:m2})
\STATE Announce $u_t$ before dealers quote
\STATE Receive the quote profile $\mathbf s_t$
\STATE Allocate: $\mathbf w_t\gets M_3(u_t,\mathbf s_t)$ \hfill Eq.~(\ref{eq:m3})
\STATE Route the $Q$ customer events by $\mathbf w_t$ and log the trace
\ENDFOR
\end{algorithmic}
\end{algorithm}

\subsection{Quote-to-Flow Allocator: How to Route Flow}
\label{sec:allocator}

The Quote-to-Flow Allocator makes the announced advantage reach whichever dealer quotes tightest, without changing how much customers trade in total. After dealers submit $\mathbf s_t$, it converts quote competitiveness into a soft allocation target,
\begin{equation}
\pi_{i,t}=\frac{\exp[-\kappa(s_{i,t}-s_t^{\min})]}{\sum_j\exp[-\kappa(s_{j,t}-s_t^{\min})]},\quad w_{i,t}=(1-\phi u_t)\frac1N+\phi u_t\pi_{i,t}.
\label{eq:m3}
\end{equation}
At most a fraction $\phi u_t$ of customer flow follows quotes, and $\sum_iw_{i,t}=1$ keeps total demand fixed. In the primary setting, a dealer that undercuts nine rivals by three ticks during an intervention receives roughly half of the flow instead of one tenth; without intervention, allocation is uniform. Transfer cells add an incumbent quote-responsive share $\eta_0$, a fraction of flow that follows quotes even without intervention,
\begin{equation}
w^{(\eta_0)}_{i,t}=\frac{1-\eta_0-\phi u_t}{N}+(\eta_0+\phi u_t)\pi_{i,t},\qquad 0\leq\eta_0+\phi u_t\leq1.
\label{eq:eta}
\end{equation}
The primary setting uses $\eta_0=0$, so without intervention a tighter quote earns no extra flow; this isolates the allocation channel but weakens ordinary price competition by construction, an assumption Section~\ref{sec:transfer} relaxes with $\eta_0>0$. Two further variants change only the implementation: \emph{no routing} keeps the announcement but allocates uniformly, and \emph{hard-best allocation} sends all quote-responsive flow to the single tightest quote, testing whether the placement gain depends on the soft allocation rule.

\begin{table*}[t]
\caption{Experimental settings, frozen before outcome evaluation. Primary setting: $K=20$ populations, $R=10$ common paths.}
\label{tab:config}
\centering
\footnotesize
\tablestyle
\setlength{\tabcolsep}{4pt}
\begin{tabular}{>{\raggedright\arraybackslash\leavevmode\color{tabsub}}p{62pt}>{\raggedright\arraybackslash}p{165pt}@{\hspace{14pt}}>{\raggedright\arraybackslash\leavevmode\color{tabsub}}p{68pt}>{\raggedright\arraybackslash}p{173.3pt}}
\toprule
\rowcolor{tabband}
\multicolumn{2}{l}{\textbf{A}\enspace\textbf{Market environment}} & \multicolumn{2}{l}{\textbf{C}\enspace\textbf{Learners}} \\
Market & $N=10$ dealers, $T=100$ decisions, $Q=20$ unit events per decision; half-spreads $\mathcal S=\{1,\ldots,7\}$ ticks & Controlled response & Shared $2\times64$ tanh actor, dealer-specific heads, linear response adapter, centralized critic \\
Risk process & Shock scales $\sigma_{\mathrm L}/\sigma_{\mathrm H}=0.75/4.00$; transitions $p_{\mathrm{LH}}/p_{\mathrm{HL}}=0.04/0.10$; $p^0_{\mathrm H}=0.30$ & Controlled PPO & Learning rate $3\times10^{-4}$, entropy $0.02$, $\gamma=0.96$, GAE $\lambda=0.95$, clip $0.20$, 4 epochs, batch 512 \\
Order flow & Informed probability $0.80$; inventory penalty $0.002$ & Full-policy learners & Recurrent MAPPO/IPPO, hidden size 64, complete actor--critic updates; learning rate $5\times10^{-4}$, entropy $0.01$, $\gamma=0.99$, GAE $\lambda=0.95$, clip $0.20$, 10 epochs \\
\rowcolor{tabband}
\multicolumn{2}{l}{\textbf{B}\enspace\textbf{Primary contract}} & \multicolumn{2}{l}{\textbf{D}\enspace\textbf{Protocol and inference}} \\
State Scorer ($M_1$) & 10-step volatility and absolute return, fit on 1,000 untreated records (validation AUC 0.83 on 1,000 more) & Episodes & 384 untreated, then 96 arm-specific response episodes; primary $K=20$, $R=10$; transfer cells $K=10$ (hard-best, stress test $K=20$), $R=10$ \\
Scheduler ($M_2$) & Threshold $\tau=0.681$, pacing $\rho=0.35$, $B_{\max}=2$ & Inference & Student-$t$ intervals across populations (primary), exact sign tests, paired two-way bootstrap; Holm-adjusted link contrasts \\
Allocator ($M_3$) & Quote-aligned allocation, $\kappa=4$, $\phi=0.50$, $\eta_0=0$ & & \\
\bottomrule
\end{tabular}
\end{table*}

\subsection{Evaluation Protocol}
\label{sec:protocol}

The paired re-adaptation protocol evaluates each link after dealers adapt. On each common market path, the placement arms receive the same score-derived strength sequence, and a fixed permutation changes only its timing (Figure~\ref{fig:teaser}b). Any difference between placement arms therefore reflects where the strengths land, not how many or how strong they are. Every arm then starts from the same population checkpoint and re-adapts independently with the same training budget; common market paths and random numbers reduce evaluation noise. Matching fixes the announced amount of intervention but not the realized routing shift, which also depends on learned quotes, so we report routing intensity (routing total variation) as a diagnostic.

The stale-announcement control shows dealers $u_{t-1}$ while the venue applies $u_t$, testing whether it matters that dealers see the current commitment. The identity-scrambling control reassigns the quote-derived weights across dealers with a fixed non-identity permutation: the set of weights, targeted quantity, maximum share, routing intensity, and total demand are unchanged, but a dealer's flow no longer follows its own quote. Because each arm induces a different learned response, the effects are informative but not additive. Every quantity from market state to outcome is logged, so each comparison can be recomputed from the traces.

\section{Experiments}
\label{sec:experiments}

\subsection{Experimental Setup}

We evaluate four questions. \textbf{RQ1: State placement (CH1).} Does giving the same set of strengths to higher-scored states lower low-risk execution cost after dealers adapt? \textbf{RQ2: Pre-quote information (CH2).} Does seeing the current, rather than the previous, announcement before quoting help? \textbf{RQ3: Quote-to-flow implementation (CH3).} Does it matter that the flow advantage follows the dealer's own quote? \textbf{RQ4: Robustness.} How sensitive are these effects to pacing, allocation, the choice of learner, dealer count, and incumbent quote responsiveness?

We implement the study in ABIDES, an event-driven, message-level market simulator with reinforcement-learning interfaces \cite{byrd2020abides,amrouni2021abides,cheridito2025abides}; dealer reward follows inventory-aware market-making formulations \cite{ho1981optimal,avellaneda2008high,gueant2013dealing,spooner2020robust}. We use PPO with generalized advantage estimation \cite{schulman2015high,schulman2017proximal}, with recurrent MAPPO and IPPO as centralized and independent learners \cite{yu2022surprising,de2020independent}.

The comparison arms map to these questions: exact random, circular-shift, and periodic schedules are budget-matched placement controls (RQ1); the one-step-stale announcement is the information control (RQ2); identity scrambling and no routing are implementation controls (RQ3); and hard-best allocation, MAPPO and IPPO learners, other dealer counts, nonzero $\eta_0$, and a wider-action stress test probe robustness (RQ4). No intervention and a calibrated lagged-spread trigger serve as practical reference points. Frozen adaptations of a Brero-style buy-box rule and a Banerjee-style commitment rule \cite{brero2022learning,banerjee2026interpretingcounteringcollusiondeeplearning} did not pass the prespecified gate on stability, activation, and routing intensity, so they are not ranked against these controls.

Because deep-RL and MARL results are seed-sensitive \cite{henderson2018deep,agarwal2021deep,papoudakis2020benchmarking,gorsane2022towards,patterson2024empirical}, independently trained populations, not repeated evaluation paths, are the unit of inference. The primary setting has 20 populations, each evaluated on 10 common paths, and paired path differences are averaged within each population. We report two-sided 95\% Student-$t$ intervals across populations, check them with exact sign tests and a paired population/path bootstrap, and Holm-adjust the link contrasts. Table~\ref{tab:config} lists the frozen settings.

For each population, one checkpoint is first trained without intervention. Every arm starts from this checkpoint, freezes the base policy, and trains a zero-initialized, dealer-specific response adapter with a centralized critic for a fixed number of episodes; dealers observe the current $u_t$, or $u_{t-1}$ in the stale arm. We call this restricted learner the \emph{controlled response}: dealers adjust their quotes to each rule while their base policy stays fixed. Full-policy learners (MAPPO and IPPO) instead update complete recurrent policies and critics under each arm. Arms of a population therefore share initialization, evaluation paths, and training budget, and differ only in the intervention rule that they face.

\begin{figure*}[t]
\centering
\includegraphics{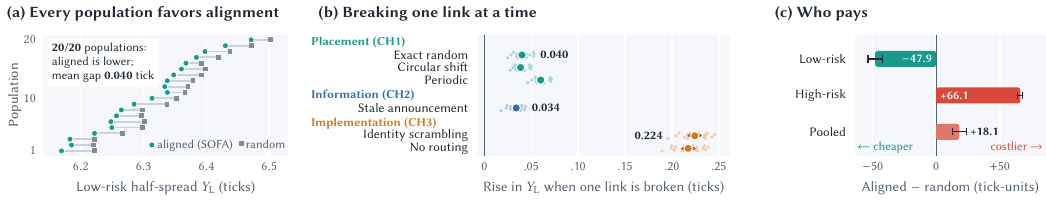}
\caption{Primary results across $K{=}20$ populations. (a) One row per population; alignment wins in all 20. (b) Small dots: populations; large: means with 95\% CIs, colored by link. (c) Low-risk customer cost falls; high-risk and pooled costs rise.}
\label{fig:main-results}
\Description{Panel a is a dot plot with one row per population, connecting the aligned and random low-risk half-spreads; the aligned value is lower in every row. Panel b shows, for six controls grouped and colored by placement, information, and implementation, each population's increase in the low-risk half-spread and the mean with its confidence interval; identity scrambling and no routing are far to the right. Panel c shows horizontal bars of the change in customer cost: negative for low-risk states and positive for high-risk and pooled states.}
\end{figure*}

\begin{table*}[t]
\caption{State alignment improves the targeted state but shifts customer cost toward high-risk states: low-risk cost falls in 20/20 populations, high-risk and pooled costs rise (Panel B), and neither flow concentration nor routing intensity increases (Panel C).}
\label{tab:main-results}
\centering
\footnotesize
\tablestyle
\setlength{\tabcolsep}{4pt}
\begin{tabular}{p{162.3pt}
  S[table-format=1.3, table-column-width=52pt]
  S[table-format=1.3, table-column-width=52pt]
  S[table-format=2.2, table-column-width=66pt]
  S[table-format=+1.3(1), table-column-width=74pt]
  >{\centering\arraybackslash}p{52pt}}
\toprule
\rowcolor{tabband}
\textbf{A}\enspace\textbf{Budget-matched schedules} & {$Y_{\mathrm L}$ (ticks)\,$\downarrow$} & {$Y_{\mathrm{all}}$ (ticks)\,$\downarrow$} & {High-risk share (\%)\,$\downarrow$} & {Increase vs.\ aligned} & Aligned better \\
\midrule
\rowcolor{tabours}
State-aligned (ours) & \bfseries 6.307 & 6.351 & \bfseries 9.42 & {--} & -- \\
Exact random & 6.347 & 6.342 & 19.44 & +0.040 \pm 0.004 & 20/20 \\
Circular shift & 6.346 & \bfseries 6.341 & 21.54 & +0.039 \pm 0.003 & 20/20 \\
Periodic placement & 6.367 & 6.347 & 21.61 & +0.060 \pm 0.003 & 20/20 \\
\end{tabular}\par\nointerlineskip
\begin{tabular}{p{86.3pt}
  S[table-format=1.3, table-column-width=40pt]
  S[table-format=1.3, table-column-width=40pt]
  S[table-format=+1.3(1), table-column-width=66pt]
  S[table-format=5.2, table-column-width=46pt]
  S[table-format=5.2, table-column-width=46pt]
  S[table-format=+2.2(3), table-column-width=66pt]
  >{\centering\arraybackslash}p{52pt}}
\midrule
\rowcolor{tabband}
\textbf{B}\enspace\textbf{State incidence} & \multicolumn{3}{c}{Execution half-spread (ticks)\,$\downarrow$} & \multicolumn{3}{c}{Customer cost (tick-units)\,$\downarrow$} & \\
\arrayrulecolor{tabhair}\cmidrule(lr){2-4}\cmidrule(lr){5-7}\arrayrulecolor{black}
\rowcolor{tabband}
State & {Aligned} & {Random} & {Change} & {Aligned} & {Random} & {Change} & Aligned better \\
\midrule
Low-risk (target) & 6.307 & 6.347 & -0.040 \pm 0.004 & 9148.97 & 9196.89 & -47.92 \pm 5.61 & 20/20 \\
High-risk & 6.440 & 6.309 & +0.131 \pm 0.007 & 3553.14 & 3487.08 & +66.06 \pm 1.96 & 0/20 \\
Pooled & 6.351 & 6.342 & +0.009 \pm 0.003 & 12702.11 & 12683.97 & +18.14 \pm 5.42 & 2/20 \\
\end{tabular}\par\nointerlineskip
\begin{tabular}{p{168.3pt}
  S[table-format=1.4, table-column-width=70pt]
  S[table-format=1.4, table-column-width=70pt]
  S[table-format=+1.4(2), table-column-width=100pt]
  >{\centering\arraybackslash}p{58pt}}
\midrule
\rowcolor{tabband}
\textbf{C}\enspace\textbf{Dealer and market-structure guardrails} & {Aligned} & {Random} & {Change} & Aligned better \\
\midrule
Aggregate dealer reward ($\times10^{3}$ units) & 9.378 & 9.339 & +0.039 \pm 0.007 & 20/20 \\
Inventory dispersion (units)\,$\downarrow$ & 8.897 & 9.422 & -0.525 \pm 0.092 & 20/20 \\
Routing intensity (TV) & 0.0622 & 0.0629 & -0.0007 \pm 0.0005 & -- \\
Normalized realized-flow HHI\,$\downarrow$ & 0.0689 & 0.0713 & -0.0024 \pm 0.0003 & 20/20 \\
Maximum realized flow share\,$\downarrow$ & 0.2509 & 0.2541 & -0.0032 \pm 0.0005 & 20/20 \\
\bottomrule
\end{tabular}
\tabnote{Means over $K=20$ populations; $\pm$ is the half-width of the 95\% Student-$t$ interval. Panel A reports each control minus the aligned arm; Panels B--C report aligned minus random. High-risk share: fraction of announced interventions that fall in high-risk states. Across populations, the Panel A levels have SD 0.080--0.084 tick. Bold: best in column.}
\end{table*}

\subsection{Main Results}
\label{sec:main-results}

Table~\ref{tab:main-results} and Figure~\ref{fig:main-results} answer RQ1. With the same intervention strengths, state alignment lowers low-risk execution cost relative to exact random placement by 0.040 tick, and all 20 populations favor alignment (Figure~\ref{fig:main-results}a); the population-level interval, the exact sign test, and the bootstrap all rule out a zero effect. Pairing is what makes this margin visible: the level of low-risk execution cost has a cross-population standard deviation about twice the effect, yet every paired difference is positive. The mechanism is visible in where commitments land: alignment places about half as many announced interventions in high-risk states as the budget-matched controls (Table~\ref{tab:main-results}, Panel A). Circular-shift and periodic placement also do worse than alignment, so neither temporal clustering nor a clock-like schedule explains the placement gain.

The margin is small per trade, under one percent of the low-risk execution cost, but it comes purely from re-timing a fixed budget rather than from intervening more. Section~\ref{sec:worth} compares it with the part of the spread that risk does not explain. At a fixed budget, placement moves cost between states.

Panels B--C of Table~\ref{tab:main-results} show this shift. Relative to exact random placement, alignment lowers the total cost paid by customers in low-risk states but raises it by more in high-risk states, so pooled customer cost rises slightly; because tick-units weight each state by its executed quantity, the two state changes add up to the pooled change. Dealer reward rises, inventory dispersion falls, flow concentration does not increase, and routing intensity is slightly lower, so the gain does not come from routing more flow. These guardrails matter: a rule that lowered customer cost by squeezing dealer profits or concentrating flow in one dealer would be fragile. The opposite signs are a central finding: a pooled endpoint alone would hide the targeted improvement, and a target-only endpoint would hide the high-risk cost. Section~\ref{sec:worth} asks when this trade is worth making.

\subsection{Ablation Study}

Table~\ref{tab:component-ablation} answers RQ2 and RQ3. A one-step-stale announcement raises low-risk execution cost by 0.034 tick, so in the tested setting customers pay less when dealers quote against the current commitment; dropping the first decision, at which both arms see a zero strength, leaves this unchanged. Because the stale arm still announces and enforces its terms, this contrast values current information but does not separately identify a commitment effect. Identity scrambling raises the cost by 0.224 tick, the largest tested degradation; with routing intensity fixed by construction, the manipulation check in Table~\ref{tab:diagnostics} confirms that the quote-to-flow link is broken. Both contrasts remain significant after Holm adjustment.

\begin{table*}[t]
\caption{Link ablations at a matched amount of intervention: every tested change raises the low-risk execution cost in every population, most under identity scrambling, the most disruptive control. Effects are total learned responses, not additive shares or a general ranking.}
\label{tab:component-ablation}
\centering
\footnotesize
\tablestyle
\setlength{\tabcolsep}{4.5pt}
\begin{tabularx}{\textwidth}{l X S[table-format=1.3(1), table-column-width=80pt] >{\centering\arraybackslash}p{62pt}}
\toprule
\rowcolor{tabband}
Variant & What is removed or changed & {Increase in $Y_{\mathrm L}$ (ticks)} & SOFA better \\
\midrule
\rowcolor{tabgroup}\multicolumn{4}{l}{\linkchip{linkplace}\textbf{Placement (CH1)}\enspace{\color{tabsub}State Scorer ($M_1$) replacements and re-timed schedules}} \\
Volatility-only score & Absolute-return feature and the two-feature score (validation AUC 0.76) & 0.036 \pm 0.003 & 20/20 \\
Absolute-return-only score & Volatility feature and the two-feature score (validation AUC 0.80) & 0.068 \pm 0.004 & 20/20 \\
Lagged-spread trigger & Public-state score replaced by an outcome-linked trigger & 0.023 \pm 0.004 & 10/10 \\
Exact random placement & State alignment removed; exact strength multiset retained & 0.040 \pm 0.004 & 20/20 \\
Circular shift & State alignment removed; temporal clustering retained & 0.039 \pm 0.003 & 20/20 \\
Periodic placement & State-dependent timing replaced by fixed cadence & 0.060 \pm 0.003 & 20/20 \\
\rowcolor{tabgroup}\multicolumn{4}{l}{\linkchip{linkinfo}\textbf{Information (CH2)}\enspace{\color{tabsub}Commitment Scheduler ($M_2$) announcement before quoting}} \\
One-step-stale announcement & Current information removed before quoting & 0.034 \pm 0.003 & 20/20 \\
\rowcolor{tabgroup}\multicolumn{4}{l}{\linkchip{linkimpl}\textbf{Implementation (CH3)}\enspace{\color{tabsub}Quote-to-Flow Allocator ($M_3$) quote-to-flow link}} \\
Identity-scrambled allocation & Quote-derived weights reassigned across dealers each step; weight multiset retained & 0.224 \pm 0.006 & 20/20 \\
\bottomrule
\end{tabularx}
\tabnote{Mean $\pm$ half-width of the 95\% Student-$t$ interval across populations. Positive values indicate degradation from the full contract; score replacements are calibrated separately and are not exact-strength feature attributions; the lagged-spread row comes from a separate ten-population comparison.}
\end{table*}

The State Scorer comparisons favor the two-feature score over the single-feature and lagged-spread alternatives. The lagged-spread trigger was calibrated to a comparable activation rate and routing intensity, so it competes on placement rather than on the amount of intervention. Because each replacement is calibrated separately, these rows compare complete scores rather than single features, and a better regime classifier does not necessarily place interventions better. Each row is the total learned response to one change, so the effects are not additive, and their sizes also reflect how disruptive each control is; they do not rank the links' importance in general.

\subsection{Sensitivity Analysis}
\label{sec:sensitivity}

\begin{figure*}[t]
    \centering
    \begin{minipage}[t]{0.535\textwidth}
        \centering
        \includegraphics{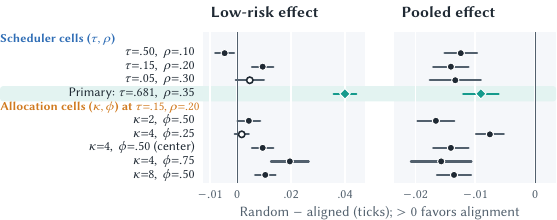}
        \caption{Placement gains are local: very low pacing reverses them, two cells are unresolved (hollow: CI includes 0), and pooled costs stay higher.}
        \label{fig:sensitivity}
        \Description{Two forest plots share row labels for four scheduler cells, including the primary setting, and five one-factor allocation cells. The left plot shows the low-risk placement effect and the right plot the pooled effect, both as random minus aligned, with population-level confidence intervals; hollow markers have intervals that include zero.}
    \end{minipage}
    \hfill
    \begin{minipage}[t]{0.43\textwidth}
        \centering
        \includegraphics{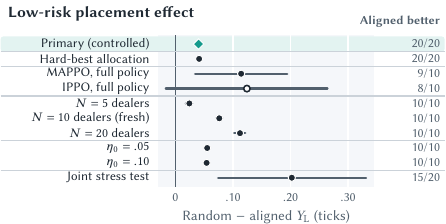}
        \caption{Transfer: the placement direction holds in every transfer cell; only IPPO is statistically unresolved.}
        \label{fig:transfer}
        \Description{A forest plot of the low-risk placement effect for the primary setting, hard-best allocation, full-policy MAPPO and IPPO, three dealer counts, two incumbent-routing shares, and the joint stress test, with population-level confidence intervals; the right margin reports how many populations favor alignment.}
    \end{minipage}
\end{figure*}

Figure~\ref{fig:sensitivity} addresses the first part of RQ4. The placement gain is local rather than monotone: very low pacing reverses it, weak quote responsiveness and the loosest scheduler cell leave it statistically unresolved, and the primary pacing and moderate-to-strong allocation cells keep it positive. The cost side is stable across this local sweep: alignment raises high-risk and pooled costs in every tested cell while flow concentration falls slightly, so the amount of intervention cannot be chosen from the targeted metric alone. The scheduler rows change the threshold and the pacing together, and the allocation rows vary one parameter at a time around a non-primary center cell, so neither group is a full factorial sweep.

A separate check on pacing points the same way: a spread-threshold controller and always-on routing inflate dealers' inventory penalties far beyond those under the paced rule (Table~\ref{tab:diagnostics}), so the evidence favors budgeted over continuous intervention.

\subsection{Robustness to Learners and Market Settings}
\label{sec:transfer}

Figure~\ref{fig:transfer} and Table~\ref{tab:learners} address the second part of RQ4. Under full-policy MAPPO, all three link effects stay positive, with larger means but wider intervals than under the controlled response, and the stale-announcement and scrambling contrasts remain significant after Holm adjustment. IPPO is positive on average but statistically unresolved. Hard-best allocation, all three dealer counts (the ten-dealer cell uses freshly trained populations at the primary dealer count), and both nonzero incumbent shares keep positive placement intervals. The direction therefore transfers across the tested cells, but the results establish neither a common effect size nor a monotone scaling law. In the dealer-count and incumbent-share cells, aligned and random arms have nearly identical routing intensity, so the realized amount of intervention stays nearly fixed in these transfers too.

\begin{table}[t]
\caption{Behavioral diagnostics: scrambling breaks the quote-to-flow link, forced undercuts reveal no detectable continuation punishment, and unpaced intervention inflates inventory penalties.}
\label{tab:diagnostics}
\centering
\footnotesize
\tablestyle
\setlength{\tabcolsep}{5pt}
\begin{tabular}{l S[table-format=+1.5(3)]}
\toprule
\rowcolor{tabband}
Diagnostic & {Estimate ($\pm$ 95\% CI)} \\
\midrule
\rowcolor{tabgroup}\multicolumn{2}{l}{\textbf{Manipulation check}\enspace{\color{tabsub}correlation of $w$ with $-s$}} \\
Quote-aligned allocation & 0.893 \\
Identity-scrambled allocation & -0.028 \\
\rowcolor{tabgroup}\multicolumn{2}{l}{\textbf{Forced undercut}\enspace{\color{tabsub}400 paired paths}} \\
High-spread recovery by $H=10$ & 0.9625 \pm 0.0184 \\
Continuation flow loss & 0.0000 \pm 0.0000 \\
Continuation reward loss & 0.0275 \pm 0.0289 \\
Rival spread compression & 0.00008 \pm 0.00010 \\
\rowcolor{tabgroup}\multicolumn{2}{l}{\textbf{Added inventory penalty}\enspace{\color{tabsub}$\times10^{2}$, versus the paced rule}} \\
35\% spread-threshold controller & 0.998 \pm 0.127 \\
Always-on routing & 2.653 \pm 0.188 \\
\bottomrule
\end{tabular}
\end{table}

\begin{table}[t]
\caption{Rise in $Y_{\mathrm L}$ (ticks; mean $\pm$ 95\% CI) when one link is broken (exact random, stale, scrambled). Under full-policy MAPPO all three mean effects stay positive; sizes do not rank links.}
\label{tab:learners}
\centering
\footnotesize
\tablestyle
\setlength{\tabcolsep}{4pt}
\begin{tabular}{l S[table-format=1.3(1), table-column-width=40pt] c S[table-format=1.3(3), table-column-width=40pt] c}
\toprule
\rowcolor{tabband}
 & \multicolumn{2}{c}{Controlled response ($K{=}20$)} & \multicolumn{2}{c}{Full-policy MAPPO ($K{=}10$)} \\
\arrayrulecolor{tabhair}\cmidrule(lr){2-3}\cmidrule(l){4-5}\arrayrulecolor{black}
\rowcolor{tabband}
Link broken & {Rise} & SOFA better & {Rise} & SOFA better \\
\midrule
\linkchip{linkplace}Placement & 0.040 \pm 0.004 & 20/20 & 0.114 \pm 0.080 & 9/10 \\
\linkchip{linkinfo}Information & 0.034 \pm 0.003 & 20/20 & 0.193 \pm 0.173 & 7/10 \\
\linkchip{linkimpl}Implementation & 0.224 \pm 0.006 & 20/20 & 0.625 \pm 0.292 & 8/10 \\
\bottomrule
\end{tabular}
\end{table}

Full-policy learners start closer to the edge of the action grid: without intervention, they quote the widest half-spread far more often than controlled-response dealers do. A prespecified joint stress test trains fresh full-policy MAPPO populations with twice as many actions and a nonzero incumbent share. It retains a positive placement effect, with most populations favoring alignment, and its learned actions reach the edges of the action grid less often than in the primary setting. Because the wider grid and the incumbent share change together, the test does not attribute robustness to either factor alone.

\subsection{Behavioral Analysis}
\label{sec:behavior}

The behavioral tests in Table~\ref{tab:diagnostics} separate the immediate allocation channel from punishment-based interpretations. In each forced-undercut path, one dealer is forced to quote below prevailing spreads for a short window, and flow, rewards, and rival quotes are then tracked. Dealers usually return to wide spreads within ten decisions, but the undercutting dealer loses no continuation flow, and the intervals for its continuation reward loss and for rival spread compression include zero. Over the tested horizon, the evidence supports an immediate quote-to-flow response and does not detect the continuation punishment that a punishment-supported interpretation would require.

\subsection{Economic Interpretation}
\label{sec:worth}

\begin{figure}[t]
\centering
\includegraphics{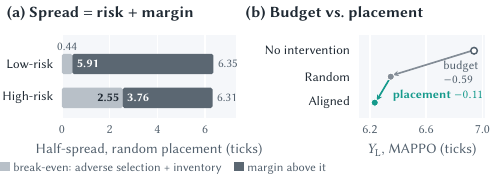}
\caption{Context for the placement margin. (a) The low-risk spread is almost all margin above break-even, whereas about 40\% of the high-risk spread is break-even. (b) Under full-policy MAPPO, the paced budget does most of the work.}
\label{fig:context}
\Description{Panel a shows two stacked horizontal bars of the half-spread under random placement. Low-risk: 0.44 tick break-even plus 5.91 ticks of margin, 6.35 in total. High-risk: 2.55 break-even plus 3.76 margin, 6.31 in total. Panel b shows the low-risk half-spread under full-policy MAPPO: 6.94 ticks with no intervention, 6.35 under random placement (a 0.59 drop from the budget), and 6.23 under aligned placement (a further 0.114 drop from placement).}
\end{figure}

\begin{table}[t]
\caption{State incidence beyond the primary setting: change in half-spread under alignment (aligned $-$ random, ticks; mean $\pm$ 95\% CI). The low-risk gain holds in every population of every cell and the high-risk cost rises in every cell, but the sign of the pooled change differs across cells.}
\label{tab:incidence}
\centering
\footnotesize
\tablestyle
\setlength{\tabcolsep}{3.5pt}
\begin{tabular}{l S[table-format=+1.3(2)] S[table-format=+1.3(2)] S[table-format=+1.3(1)]}
\toprule
\rowcolor{tabband}
Cell & {Low-risk} & {High-risk} & {Pooled} \\
\midrule
\rowcolor{tabours}
Primary ($K{=}20$) & -0.040 \pm 0.004 & +0.131 \pm 0.007 & +0.009 \pm 0.003 \\
5 dealers & -0.024 \pm 0.006 & +0.076 & +0.005 \pm 0.005 \\
10 dealers (fresh) & -0.076 \pm 0.005 & +0.218 & -0.006 \pm 0.004 \\
20 dealers & -0.112 \pm 0.010 & +0.155 & -0.018 \pm 0.006 \\
$\eta_0=0.05$ & -0.055 \pm 0.005 & +0.129 \pm 0.012 & -0.009 \pm 0.003 \\
$\eta_0=0.10$ & -0.054 \pm 0.004 & +0.137 \pm 0.009 & -0.006 \pm 0.002 \\
\bottomrule
\end{tabular}
\tabnote{Transfer cells: $K=10$ populations each; the ten-dealer cell uses fresh populations at the primary dealer count. Dealer-count cells report no high-risk intervals.}
\end{table}

In the primary setting, alignment moves customer cost between states rather than lowering it (Section~\ref{sec:main-results}), so adopting it is a choice of venue objective. A venue that weights a tick-unit of customer cost by $\omega_{\mathrm L}$ in low-risk and $\omega_{\mathrm H}$ in high-risk states prefers alignment only if $\omega_{\mathrm L}/\omega_{\mathrm H}>1.38$, the ratio of the high-risk cost increase to the low-risk saving (Table~\ref{tab:main-results}, Panel B). Equal weights, which treat every tick-unit alike, reject it. A venue whose concern grows quadratically with the part of the spread that risk does not explain accepts it. Under random placement, the excess over the break-even half-spread, which would just cover realized adverse-selection and inventory costs, is larger in low-risk than in high-risk states (Figure~\ref{fig:context}a), and a quadratic penalty on this excess gives $\omega_{\mathrm L}/\omega_{\mathrm H}\approx1.57$. Dealer reward and inventory dispersion act as participation and stability constraints rather than objectives---a rule that squeezed dealer reward or raised inventory risk would erode dealers' willingness to quote---and neither deteriorates here (Table~\ref{tab:main-results}, Panel C), although part of the higher reward is the higher pooled customer cost.

Two further comparisons put the trade in context. Against no intervention, available only under full-policy MAPPO, the paced budget itself accounts for most of the drop in low-risk execution cost, and alignment adds a smaller further drop (Figure~\ref{fig:context}b), so placement refines what a paced budget already achieves. Across transfer cells, the incidence is not fixed (Table~\ref{tab:incidence}): high-risk cost rises in every cell, but with ten (fresh) or twenty dealers, or with a nonzero incumbent share, the low-risk gain outweighs it and pooled cost falls, so even equal weights would accept alignment there; with five dealers the pooled change is unresolved. Because the fresh ten-dealer cell reverses the pooled sign of the primary setting at the same dealer count, the net incidence varies across cells rather than with dealer count alone. A low-risk classification alone does not show that the targeted margins are strategic, however: we detect no punishment (Section~\ref{sec:behavior}), and the break-even level covers only the risks that the simulator models.

\section{Limitations}
\label{sec:limitations}

Our evidence comes from a stylized ABIDES market and a specific re-adaptation procedure. Dealers are homogeneous and quote symmetric half-spreads on a coarse price grid, customer demand is fixed, price and risk paths are exogenous, and there are no strategic background traders; with elastic demand, the targeted gain could grow while liquidity in high-risk states thins. Incumbent quote-following was tested only at small shares, so venues where most flow already follows the best quote are untested. The State Scorer is trained on the simulator's regime labels, whereas a deployed venue would have to rely on noisier labels observed after the fact, such as realized volatility. The main results use a frozen base policy with a briefly trained response adapter; full-policy MAPPO agrees, with wider intervals, while IPPO remains unresolved, and we did not trace longer training or learning curves. Finally, the sensitivity analysis is local, and the effect can vanish or reverse outside the tested parameter region.

\section{Conclusion}
\label{sec:conclusion}

We studied when and how a venue-side order-flow intervention works after dealers adapt, using \textbf{SOFA} and a paired re-adaptation protocol that separates placement, information, and implementation at a fixed amount of intervention. In the tested market, state alignment yields a consistent low-risk benefit, current announcements outperform stale ones, and breaking the quote-to-flow link causes the largest tested degradation. In the primary setting the benefit shifts customer cost toward high-risk states rather than lowering it overall, although pooled cost falls in four of the five transfer cells of Table~\ref{tab:incidence}. For a venue that prioritizes low-risk execution, the results favor announcing a budgeted, state-dependent flow advantage that follows each dealer's own quote. More broadly, interventions for learning agents should be judged at a fixed amount of intervention, after re-adaptation, one link at a time; this turns a comparison of final outcomes into an account of why an intervention works and where its costs fall. We therefore see the protocol, more than SOFA's effect size, as the reusable contribution: freeze a public score, match intervention strengths, start every rule from a common checkpoint, re-adapt, break one link at a time, and infer across independently trained populations. Applying this recipe to other venue interventions, and to an externally validated public risk score, is the next test of generality.

\bibliographystyle{ACM-Reference-Format}
\bibliography{reference}

\end{document}